\documentclass[10pt,journal,compsoc]{IEEEtran}
\ifCLASSOPTIONcompsoc
  \usepackage[nocompress]{cite}
\else
  \usepackage{cite}
\fi
\ifCLASSINFOpdf
\else
\fi
\usepackage{lipsum}
\usepackage{mathtools}
\usepackage{amsthm,amsmath,amssymb}
\usepackage{mathptmx}
\usepackage{mathrsfs}
\usepackage{bm}
\usepackage{diagbox}
\usepackage{tabularx}
\usepackage{multirow}
\usepackage{bigstrut}
\usepackage{graphicx}
\usepackage{array}
\usepackage{stfloats}
\usepackage{colortbl}
\usepackage{setspace}
\usepackage{soul}
\usepackage{color}
\usepackage[table ]{ xcolor}
\begin{document}
%
\title{A Neural-symbolic Framework under Statistical Relational Learning}
%
%
%
%

\author{Dongran~Yu,~\IEEEmembership{}
        Xueyan~Liu,~\IEEEmembership{}
        Shirui~Pan,~\IEEEmembership{}
        Anchen~Li~\IEEEmembership{}
        and~Bo~Yang~\IEEEmembership{}
\IEEEcompsocitemizethanks{\IEEEcompsocthanksitem B. Yang (corresponding author), X. Liu (corresponding author) and A. Li are with the Key Laboratory of Symbolic Computation and Knowledge Engineer, Ministry of Education, Jilin University, Changchun, Jilin 130012, China and the School of Computer Science and Technology, Jilin University, Changchun, Jilin 130012, China.\protect\\
E-mail: ybo@jlu.edu.cn; xueyanliu@jlu.edu.cn; liac20@mails.jlu.edu.cn
\IEEEcompsocthanksitem D. Yu is with the Key Laboratory of Symbolic Computation and Knowledge Engineer, Ministry of Education, Jilin University, Changchun, Jilin 130012, China, and the School of Artificial Intelligence, Jilin University, Changchun, Jilin 130012, China. \protect\\
E-mail: yudran@foxmail.com \protect\\
\IEEEcompsocthanksitem S. Pan is with School of Information and Communication Technology, Griffith University, Brisbane 4222, Queensland, Australia. \protect\\
E-mail: s.pan@griffith.edu.au}
\thanks{}}

\IEEEtitleabstractindextext{%
\begin{abstract}
A key objective in the field of artificial intelligence is to develop cognitive models that can exhibit human-like intellectual capabilities. One promising approach to achieving this is through neural-symbolic systems, which combine the strengths of deep learning and symbolic reasoning. However, current methodologies in this area face limitations in integration, generalization, and interpretability. To address these challenges, we propose a neural-symbolic framework based on statistical relational learning, referred to as \texttt{NSF-SRL}. This framework effectively integrates deep learning models with symbolic reasoning in a mutually beneficial manner. In \texttt{NSF-SRL}, the results of symbolic reasoning are utilized to refine and correct the predictions made by deep learning models, while deep learning models enhance the efficiency of the symbolic reasoning process. Through extensive experiments, we demonstrate that our approach achieves high performance and exhibits effective generalization in supervised learning, weakly supervised and zero-shot learning tasks. Furthermore, we introduce a quantitative strategy to evaluate the interpretability of the model’s predictions, visualizing the corresponding logic rules that contribute to these predictions and providing insights into the reasoning process. We believe that this approach sets a new standard for neural-symbolic systems and will drive future research in the field of general artificial intelligence.
\end{abstract}

}

\maketitle

\IEEEdisplaynontitleabstractindextext

%
\IEEEpeerreviewmaketitle

\IEEEraisesectionheading{\section{Introduction}\label{sec:introduction}}

%
%
%
%
\IEEEPARstart{H}{uman} cognitive systems encompass both perception and reasoning. Specifically, perception is primarily responsible for recognizing information, while reasoning handles logical deduction and analytical thinking. When humans process information, they integrate both perception and reasoning to enhance their comprehension and decision-making capabilities. Current artificial intelligence systems typically specialize in either perception or reasoning. For instance, deep learning models excel in perception, achieving remarkable performance in tasks that involve inductive learning and computational efficiency. In contrast, symbolic logic is adept at logical reasoning, providing strong results in deductive reasoning tasks, generalization, and interpretability. However, both models have inherent limitations. Deep learning models often operate as black boxes, lacking interpretability, generalizing poorly, and requiring vast amounts of training data to perform optimally. On the other hand, symbolic logic relies on search algorithms to explore solution spaces, resulting in slow reasoning in large-scale environments. Therefore, integrating the strengths of both models offers a way to combine perception and reasoning into a unified framework that more effectively mimics human cognitive processes. As Leslie G. Valiant argues, reconciling the statistical nature of learning with the logical nature of reasoning to create cognitive computing models that integrate concept learning and manipulation is one of the three fundamental challenges in computer science \cite{valiant2003three}.

The neural-symbolic system represents a promising approach for effectively integrating perception and reasoning into a unified framework \cite{belle2020symbolic, Hitzler2021NeuroSymbolicAI, curry2022multimodal}. Various neural-symbolic systems have been proposed, which can be broadly classified into three categories \cite{yu2021recent}: learning-for-reasoning methods, reasoning-for-learning methods, and learning-reasoning methods. Learning-for-reasoning methods \cite{qu2019probabilistic, zhang2020efficient, mao2019neuro} primarily focus on symbolic reasoning. In these methods, deep learning models transform unstructured inputs into symbolic representations, which are then processed by symbolic reasoning models to derive solutions. In some cases, deep learning models replace search algorithms, thus accelerating symbolic reasoning. Reasoning-for-learning approaches \cite{xu2018semantic, xie2019embedding, luo2020context} focus more on deep learning. Symbolic knowledge is encoded into distributed representations and integrated into deep learning models to compute results. However, these methods often use deep learning to support symbolic reasoning or incorporate symbolic priors to enhance deep learning, without fully achieving complementary integration. Few studies explore learning-reasoning methods, which aim for more comprehensive integration \cite{manhaeve2021neural, zhou2019abductive}. For example, Manhaeve et al. \cite{manhaeve2018deepproblog} combine a deep learning model with a probabilistic logic programming language, where the output of the deep learning model serves as input for symbolic reasoning. Techniques like arithmetic circuits and gradient semi-rings enable interaction between the deep learning model and symbolic reasoning. Zhou \cite{zhou2019abductive} integrates machine learning with logic reasoning based on the principle of abduction, using the machine learning model’s output as input for logical reasoning. This reasoning process iteratively corrects the model's output through consistency optimization, and the refined output is then used as supervised information for further training. While these approaches represent significant progress in neural-symbolic systems, achieving full integration remains a challenging and open problem, necessitating further exploration and research.

This paper introduces a novel framework called the \texttt{N}eural \texttt{S}ymbolic \texttt{F}ramework under \texttt{S}tatistical \texttt{R}elational \texttt{L}earning  (\texttt{NSF-SRL} for short), which aims to integrate deep learning models with symbolic logic in a mutually beneficial manner. In \texttt{NSF-SRL}, symbolic logic enhances deep learning models by making their predictions more logical, consistent with common sense, and interpretable, thereby improving their generalization capabilities. In turn, deep learning enhances symbolic logic by increasing its efficiency and robustness to noise. However, a key challenge in constructing the \texttt{NSF-SRL} framework is determining \textit{how to effectively combine deep learning and symbolic logic to model a joint probability distribution.}

Statistical Relational Learning (SRL) \cite{getoor2007introduction} serves as a bridge between statistical models, such as deep learning, and relational models, like symbolic logic, by integrating the two approaches. Inspired by this framework, we employs SRL techniques to address the challenge of model construction. In this approach, deep learning processes data according to specific tasks and generates corresponding outputs, while symbolic logic learns a joint probability distribution based on these outputs and symbolic knowledge, thus constraining deep learning’s predictions to achieve mutual enhancement. It is important to note that in our framework, deep learning not only functions as a data processor for symbolic logic but also replaces traditional search algorithms to improve computational efficiency. In this study, symbolic knowledge is represented using First-Order Logic (FOL). During the training phase, the model learns the basic concepts \footnote{In this paper, concepts refer to predicates in FOL.} in FOL from the sample data, a process we term concept learning. In the testing phase, the model utilizes existing or newly acquired FOLs to combine and manipulate learned concepts, thereby generating new ones—a process referred to as concept manipulation.

In summary, our contributions can be characterized in threefold:

\begin{itemize}
\item[$\bullet$] 
In this study, we propose a general neural-symbolic system framework \texttt{NSF-SRL} and develop an end-to-end model. 
\end{itemize}

\begin{itemize}
\item[$\bullet$] 
The model employs statistical relational learning  techniques to integrate deep learning and symbolic logic, thereby achieving mutual enhancement of learning and reasoning. This integration improves the model's generalization ability and interpretability.
\end{itemize}
 
\begin{itemize}
\item[$\bullet$] 
Based on our experimental results, we demonstrate that \texttt{NSF-SRL} outperforms comparable methods in various reasoning tasks, including supervised, weakly supervised, and zero-shot learning scenarios, with respect to performance and generalization. Additionally, we emphasize the interpretability of our model by providing visualizations that enhance the understanding of the reasoning process.
\end{itemize}

In our previous conference paper \cite{yu2022probabilistic}, we initially presented and validated the proposed approach for visual relationship detection. However, this current study significantly extends that work by introducing new model designs, such as concept manipulation, incorporating new tasks like digit image addition and zero-shot image classification, and comparing against additional baseline approaches. Furthermore, we provide extensive experimental validations and comparisons to thoroughly evaluate the model's performance.
\section{Related work}
\noindent \textbf{Neural-symbolic systems}. In recent times, neural-symbolic reasoning has gained significant attention and can be classified into three main groups \cite{yu2021recent}. The first group consists of methods where deep neural networks assist symbolic reasoning. These methods replace traditional search algorithms in symbolic reasoning with deep neural networks to reduce the search space and improve computation speed \cite{qu2019probabilistic, zhang2020efficient, mao2019neuro, abboud2020learning}. For example, Qu et al. \cite{qu2019probabilistic} proposed probabilistic Logic Neural Networks (pLogicNet), which addresses the problem of reasoning in knowledge graphs (triplet completion) as an inference problem involving hidden variables in a probabilistic graph model. The pLogicNet employs a combination of variational EM and neural networks to approximate the inference. Building on the idea of pLogicNet, Zhang et al. \cite{zhang2020efficient} introduced ExpressGNN, which leverages Graph Neural Networks (GNNs) as approximate inference methods for posterior calculation in the variable EM algorithm. Marra et al. \cite{marra2021neural} proposed NMLN, which reparametrizes the MLN through a neural network that is evaluated based on input features. The second group focuses on symbolic reasoning aiding deep learning models during the learning process. These methods incorporate symbolic knowledge into the training of deep learning models to enhance performance and interpretability \cite{xie2019embedding, luo2020context, hu2016harnessing, sun2020neural, oltramari2021generalizable}. Symbolic knowledge is often used as a regularizer during training. For instance, Xie et al. \cite{xie2019embedding} encode symbolic knowledge into neural networks by designing a regularization term in the loss function for a specific task. The third group consists of models that strike a balance between deep learning models and symbolic reasoning, allowing both paradigms to contribute to problem-solving \cite{manhaeve2021neural, zhou2019abductive, badreddine2022logic, tian2022weakly}. Zhou \cite{zhou2019abductive} establishes a connection between machine learning and symbolic reasoning frameworks based on the characteristics of symbolic reasoning, such as abduction. Duan et al. \cite{duan2022deeplogic} proposed a framework for joint learning of neural perception and logical reasoning, where the two components are mutually supervised and jointly optimized. Pryor et al. \cite{pryor2023neupsl} introduced NeuPSL, where the neural network learns the predicates for logical reasoning, while logical reasoning imposes constraints on the neural network. Yang et al., \cite{yang2021neurasp} proposed NeurASP, which leverages a pre-trained neural network in symbolic computation and enhances the neural network’s performance by applying symbolic reasoning. In contrast to the aforementioned methods, our approach takes a different route to bridge the gap between deep learning models and symbolic logic through statistical relational learning. By leveraging statistical relational learning, our method retains the full capabilities of both probabilistic reasoning and deep learning, offering a unique and powerful integration of the two paradigms.

\noindent \textbf{Markov Logic Networks}. To handle complexity and uncertainty of the real world, intelligent systems require a unified representation that combines first-order logic (FOL) and probabilistic graphical models. Markov Logic Networks (MLNs) achieve this by providing a unified framework that combines FOL and probabilistic graphical models into a single representation. MLN has been extensively studied and proven effective in various reasoning tasks, including knowledge graph reasoning \cite{qu2019probabilistic, zhang2020efficient}, semantic parsing \cite{tran2008event, poon2009unsupervised}, and social network analysis \cite{zhang2014identifying}. MLN is capable of capturing complexity and uncertainty inherent in relational data. However, performing inference and learning in MLN can be computationally expensive due to the exponential cost of constructing the ground MLN and NP-complete optimization problem. This limitation hinders the practical application of MLN in large-scale scenarios. To address these challenges, many works have been proposed to improve accuracy and efficiency of MLN. For instance, some studies \cite{singla2005discriminative, mihalkova2007bottom} have focused on enhancing the accuracy of MLN, while others \cite{qu2019probabilistic, zhang2020efficient, singla2006memory, khot2011learning, bach2015hinge} have aimed to improve its efficiency. In particular, two studies \cite{qu2019probabilistic, zhang2020efficient} have replaced traditional inference algorithms in MLN with neural networks. By leveraging neural networks, these approaches offer a more efficient alternative for performing inference in MLN. This integration of neural networks and MLN allows for more scalable and effective reasoning in large-scale applications.

\begin{figure*}[h]%
\centering
\includegraphics[width=\textwidth]{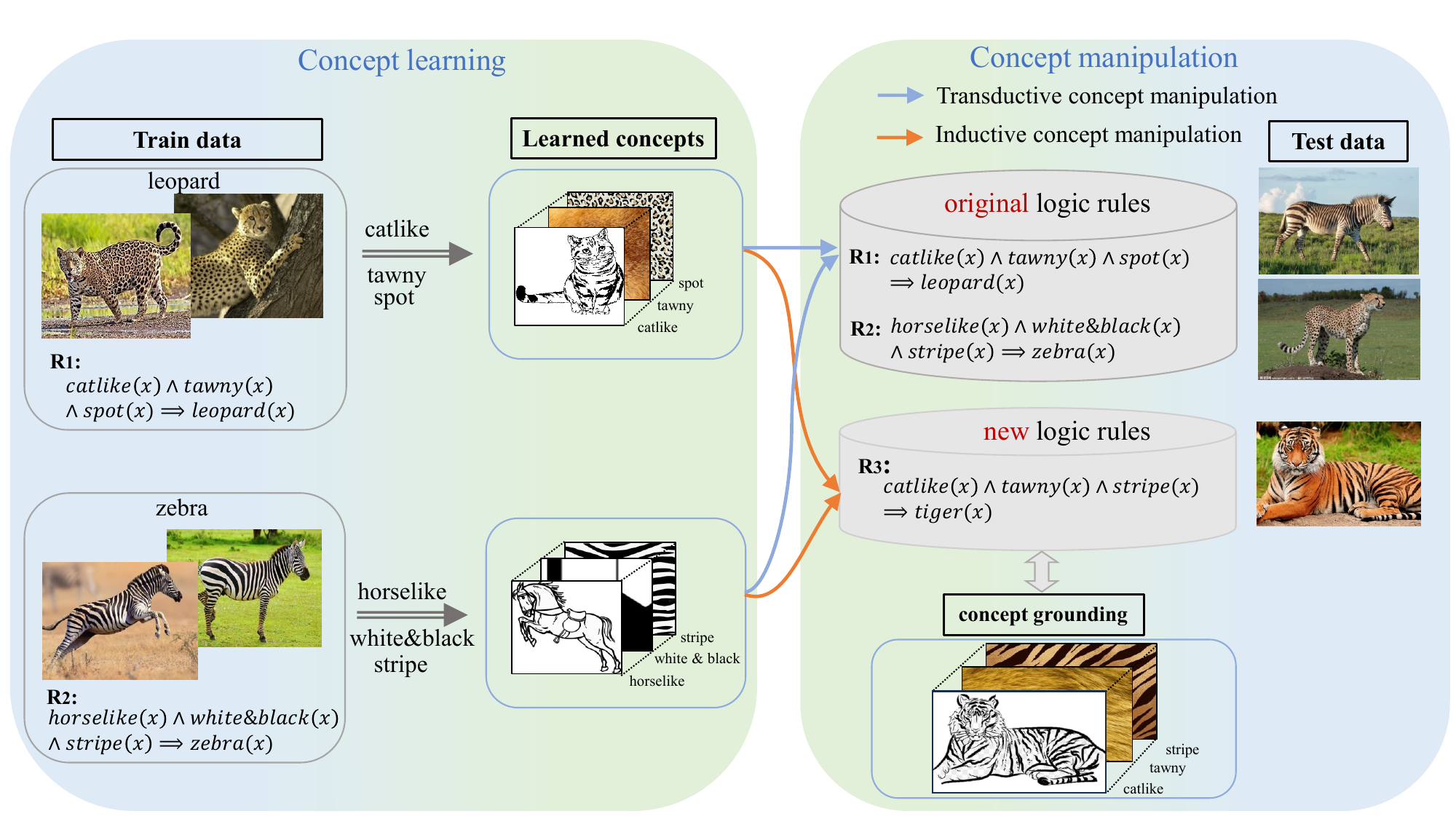}
\caption{Overview of \texttt{NSF-SRL}. The concept learning phase acquires basic concepts such as ``catlike", ``tawny" and ``spot" from the training data. In transductive concept manipulation, the learned concepts and toriginal rules are applied to test data whose labels were present in the training sets. This integration of learned concepts enhances the interpretability of \texttt{NSF-SRL} by providing insights into how predictions are made based on these concepts and the accompanying rules. Conversely, in inductive concept manipulation, the learned concepts serve as the rule body, and new rules are introduced to reason about samples with labels that have never appeared in the training set.
}\label{overview}
\end{figure*}

\section{Preliminaries}
\label{sec3}
In this section, we first introduce the neural-symbolic model definition and notations in this paper. Then, we will introduce the basic knowledge about statistic relational learning. 

\subsection{Model Description}
The primary task in developing the model \texttt{NSF-SRL} is to formulate and maximize the posterior probability $P(Y\vert X,R;\theta_1,\theta_2,w)$, where $X=\{x_1,x_2,…,x_n\}$ represents the observed data, $Y=\{y_1,y_2,…,y_n\}$ is the label set corresponding to  data $X$, and $R=\{r_1,r_2,…,r_m\}$ is the first-order logic rule set, $\theta_1$, $\theta_2$ and $w$ denote the parameters of \texttt{NSF-SRL}. $n$ is the number of the instance of raw data, and $m$ is the number of rules. Given the training dataset $D = \{(x_1,y_1 ),(x_2,y_2 ),…,(x_n,y_n)\}$ and the first-order logic rules $R$, the learning process of \texttt{NSF-SRL} can be expressed as maximizing the posterior probability, formally defined as: 
\begin{equation}\label{eq:5}
\small
\forall D \max_{\theta_1,\theta_2, w} P(Y\vert X, R;\theta_1,\theta_2,w),
\end{equation}
For example, in image classification tasks, the input data $D$ represents images, while the output $y$ corresponds to the labels of the objects within those images. To enhance understanding of this paper, symbolic descriptions are provided in Table  \ref{tab:symbol}. These descriptions clarify the symbolic representations used throughout the study and facilitate comprehension of the concepts and methodologies discussed.

\begin{table}[htpb]
\renewcommand{\arraystretch}{1.3}
\centering
\caption{Important notations and their descriptions.}
\label{tab:symbol}
\begin{tabular}{c|c}
\hline
Notations&\ Descriptions
\\
\hline
$D$&Set of input data \\
\hline
$Y$& Set of ground truths\\
\hline
$\hat{y}$& Pseudo-label\\
\hline
$R$&Set of logical rules\\
\hline
$r$& A logic rule\\
\hline
$T_r$& Triggered logic rule\\
\hline
$A$&Ground atom sets in knowledge base\\
\hline
$A_r$&Ground atom sets in a logic rule\\
\hline
$a_r$&A ground atom\\
\hline
$\phi$& Potential function\\
\hline
$\theta_1$&Parameters of neural reasoning module\\
\hline
$\theta_2$&Parameters of concept network\\
\hline
$w$&Weight sets of the logic rules \\
\hline
$w_r$& Weight of a logic rule \\
\hline
\end{tabular}

\end{table}

\subsection{Statistical Relational Learning}
Many tasks in real-world application domains are characterized by the presence of both uncertainty and complex relational structures. Statistical learning addresses the former, while relational learning focuses on the latter. Statistical Relational Learning (SRL) aims to harness the strengths of both approaches \cite{getoor2007introduction}.

In this study, we leverage SRL to integrate first-order logic (FOL, \textsf{rule body $\Rightarrow$ rule head}) with probabilistic graphical models, creating a unified framework that facilitates probabilistic inference for reasoning problems. FOL represents a type of commonsense (symbolic) knowledge that is easily understood by humans. In this paper, we treat the FOL language as a means to describe knowledge in the form of logic rules, which provides strong expressive capability \cite{enderton2001mathematical}. For instance, FOL allows for the definition of predicates and the description of various relations. 

To achieve this integration, we employ Markov Logic Networks (MLNs), a well-known statistical relational learning model, to represent FOL as undirected graphs. In the constructed undirected graph, nodes are generated based on all ground atoms \footnote{Ground atom is a replacement of all of its arguments by constants. In this paper, we refer to the process of replacement as ``grounding".}, which are logical predicates with their arguments replaced by specific constants. In this paper, $a_{r}$ denotes assignments of variables to the arguments of an FOL $r$, and all consistent assignments are captured in the set $A_{r}=\{a_{r}^{1}, a_{r}^{2},...\}$.  For instance, if we have a constant set $C=\{c_{1}, c_{2}\}$ and an FOL $r \in R$ such as $\texttt{catlike}(x) \wedge \texttt{tawny}(x) \wedge \texttt{spot}(x) \Rightarrow \texttt{leopard}(x)$, the corresponding ground atoms $A_{r}$ can be generated such as $\{\texttt{catlike}(c_{1}), \texttt{catlike}(c_{2}), \texttt{tawny}(c_{1}), \texttt{tawny}(c_{2}), \texttt{spot}(c_{1})\\, \texttt{spot}(c_{2}),\texttt{leopard}(c_{1}), \texttt{leopard}(c_{2})\}$. Furthermore, an edge is established between two nodes if the corresponding ground atoms co-occur in at least one ground FOL in the MLN. Consequently, a ground MLN can be formulated as a joint probability distribution, capturing the dependencies and correlations among the ground atoms. This joint probability distribution is expressed as:
\begin{equation}\label{eq:3}
\small
   P(A)=\frac 1{Z(w)} \exp\{\sum_{r\in R} w_r \sum_{a_{r}\in A_{r}} \phi(a_{r})\},
\end{equation}
where $Z(w) = \sum_{A} \sum_{A_r \in A, a_r \in A_r} \phi(a_{r})$ is the partition function that sums over all ground atoms. $A$ represents all ground atoms in the knowledge base, while $\phi$ is a potential function reflecting the number of times a FOL statement is true. The variable $w$ denotes the weight sets of all FOLs, and $w_{r}$ refers to the weight of a specific FOL. 

\begin{figure*}[htpb]
\centering
\includegraphics[width=1\textwidth]{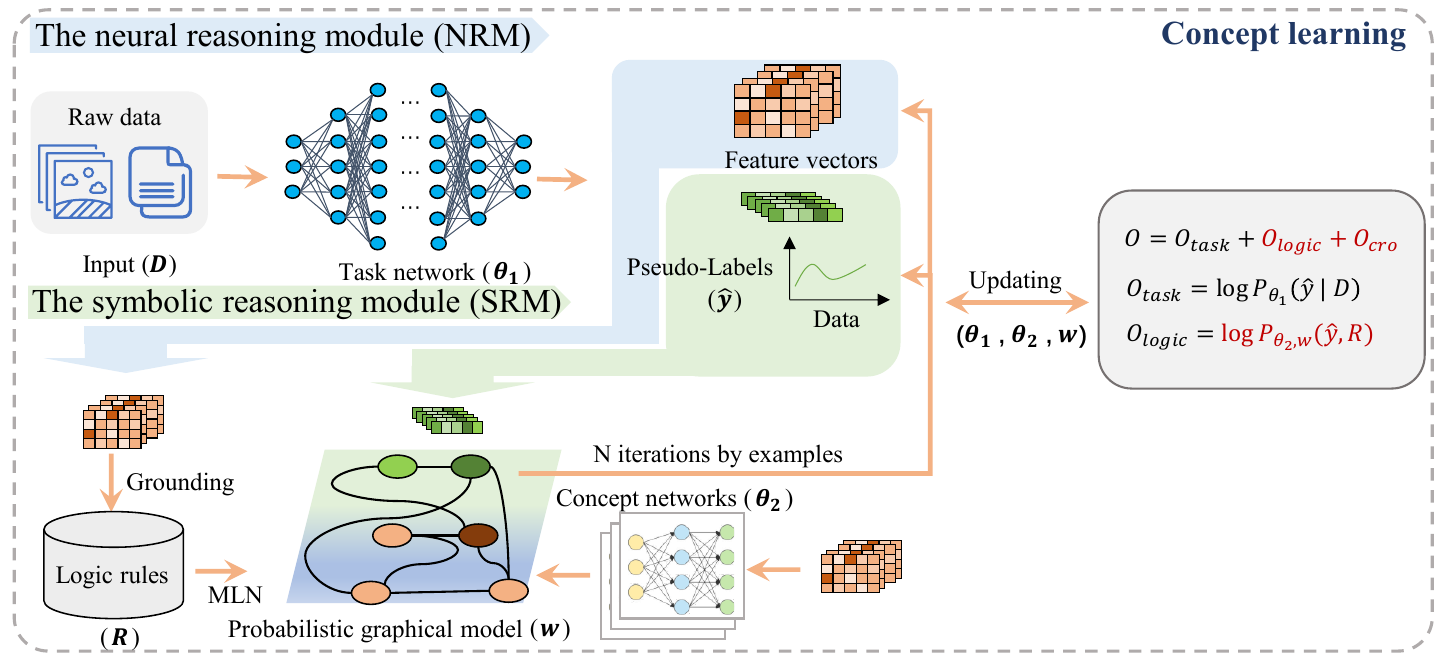}
\caption{Illustration of concept learning. The NRM aims to predict labels for raw data, generating pseudo-labels and feature vectors as outputs. The SRM is a probabilistic graphical model that incorporates both the pseudo-labels from the NRM and the ground atoms from the MLN. The entire model is trained end-to-end, using backpropagation to iteratively refine the pseudo-labels.}\label{concept_learning}
\end{figure*}

\section{Our method: NSF-SRL}
The goal of the \texttt{NSF-SRL} framework is to achieve a mutual integration of deep learning and symbolic logic. In this framework, deep learning can take the form of any task-related neural network, primarily responsible for feature extraction and result prediction. Symbolic logic, on the other hand, is grounded in probabilistic graphical models and is responsible for logical reasoning. In this section, we first provide an overview of our \texttt{NSF-SRL} in Section \ref{sec4:1}. We then present concept learning in Section \ref{sec4:2}, followed by a description of concept manipulation in Section \ref{sec4:3}.
\subsection{Overview of NSF-SRL}
\label{sec4:1}
\noindent An overview of the \texttt{NSF-SRL} framework, consisting of two key phases—concept learning and concept manipulation—is presented in Fig. \ref{overview}.

Concept learning focuses on acquiring fundamental concepts from training data. For instance, we can learn essential concepts such as ``catlike", ``tawny" and ``spot" from images of leopards, and ``horselike", ``white\&black" and ``stripe" from images of zebras, utilizing the rules R1: $\textsf{catlike}(x)\land \textsf{tawny}(x) \land \textsf{spot}(x) \Rightarrow \textsf{leopard}(x)$ and R2: $\textsf{horselike}(x) \wedge \textsf{white\&black}(x) \wedge \textsf{stripe}(x) \Rightarrow \textsf{zebra}(x)$. 

Concept manipulation is used for reasoning and interpreting results, employing existing or newly acquired symbolic knowledge to combine established concepts and generate new ones. In this paper, we identify two types of conceptual operations: transductive concept manipulation and inductive concept manipulation. In transductive concept manipulation, the learned concepts and original rules are utilized to test data whose labels have appeared in the training set. Incorporating these learned concepts enhances the interpretability of the \texttt{NSF-SRL}, providing insights into how prediction results are derived in conjunction with the rules. For example, the predicted label ``leopard" can be attributed to rule R1. Conversely, in inductive concept manipulation, the learned concepts and the new rules are applied to test data whose label has never appeared in the training set. Specifically, the learned concepts serve as the rule body of a new rule, which is used to reason the rule head as the output when testing a new sample. For instance, when an image containing a tiger is fed into the well-trained model, it can trigger the new rule R3 and generate corresponding ground atoms such as ``catlike", ``tawny" and ``stripe" via concept grounding. By leveraging R3 and the ground atoms, the model infers the new concept ``tiger". Inductive concept manipulation enables the application of previously learned concepts to new tasks, facilitating the generation of new concepts through inference and realizing adaptation and generalization to new tasks. In summary, through the process of concept manipulation, the \texttt{NSF-SRL} effectively learns, reasons, and produces explainable results by leveraging learned concepts.

\subsection{Concept Learning}
\label{sec4:2}
Concept learning involves a Neural Reasoning Module (NRM) and a Symbolic Reasoning Module (SRM), as illustrated in Fig. \ref{concept_learning}. These two modules engage in end-to-end joint learning to produce a trained model. Specifically, the NRM functions as a task network, generating pseudo-labels and feature vectors. In contrast, the SRM operates as a probabilistic graphical model responsible for deriving reasoning outcomes. During the training process, the SRM constrains the parameter learning of the NRM, enhancing the accuracy and interpretability of its predictions. After $N$ iterations and corresponding parameter updates, the trained model is achieved.

\subsubsection{Neural Reasoning Module}
The Neural Reasoning Module (NRM) is a versatile deep neural network whose architecture can vary according to the specific task at hand. This adaptability enables the NRM to accommodate diverse tasks and to be implemented with various network architectures. For instance, in the digital image addition task, the NRM may utilize a Convolutional Neural Network (CNN) to process image data, whereas in object detection, it may adopt a network structure incorporating ResNet to enhance detection performance. This capability to dynamically adjust the network architecture based on task requirements allows the NRM to effectively meet the needs of different applications. The objective to be maximized in terms of log-likelihood is formalized as follows:

\begin{equation}\label{eq:2}
\small
O_{task}=\log P_{\theta_1} (\hat y \vert D),
\end{equation}
where $\theta_1$ is the learnable parameter of the NRM. At the beginning of the model training, the NRM may produce predictions with substantial errors due to insufficient training. Consequently, in this paper, we refer to these predictions as pseudo-labels $\hat{y}$.

\subsubsection{Symbolic Reasoning Module}
The Symbolic Reasoning Module (SRM) plays a critical role in supporting the NRM by facilitating learning and employing reasoning to generate predictive outcomes and provide evidence for result interpretation. Specifically, the SRM operates as follows: when presented with a training sample $(x_i,y_i)$, it is responsible for deducing the outcome $y_i$ based on the predicted label $\hat{y_i} $, the feature vector output by the NRM, and first-order logic rules. If $\hat{y_i}$ is incorrect, the SRM adjusts the NRM parameters through backpropagation to correct the prediction. To achieve this, we leverage SRL to construct a probabilistic graphical model within the SRM, as depicted in Fig. \ref{concept_learning}.
The primary objective of the SRM is to utilize SRLs for learning variables and guiding the NRM's reasoning in the correct direction, effectively serving as an error corrector. In this study, the probabilistic graphical model is instantiated using a MLN that encompasses all tasks discussed in the validations.

When using MLNs to model logical rules, various structures can be adopted depending on the task, including single-layer and double-layer configurations. For instance, in the case of Visual Relationship Detection (VRD), we employed a double-layer structure, as detailed in Section 5.4 and illustrated in Fig. \ref{vrd_symbolic}. In other scenarios, we utilized a single-layer structure, with its joint probability distribution taking the form presented in Eq. (\ref{eq:3}). However, if the MLN incorporates multiple types of nodes and potential functions, the joint probability distribution will consist of multiple components. In this study, obtaining the nodes of the MLN requires performing grounding of the FOL statements. Grounding all FOLs in the database can lead to an excessively large number of variables, significantly increasing model complexity. Therefore, during training, the model identifies FOLs that are strongly related to the data, such as predicates that share the same labels as the data in a FOL. The optimization goal of the SRM is defined as $O_{logic}$ in Eq. (\ref{eq:4}), which aims to maximize the joint probability distribution over all variables in terms of log-likelihood,

\begin{equation}\label{eq:4}
\small
\begin{split}
O_{logic}=\log P_{\theta_{2},w}(\hat{y}, R)
=\log\{\frac{1}{Z(w)}\exp\{ \sum_{r\in R} w_r \sum_{a_{r}\in A_{r}} \phi(a_{r}) + \mathbb{C}\}\},
\end{split}
\end{equation}
where $\mathbb{C}$ represents a custom term that may include potential functions 
$\phi_{1},\phi_{2},...$, and should be designed according to task requirements.

\begin{figure}[h]
  \centering
   \includegraphics[width=1\linewidth]{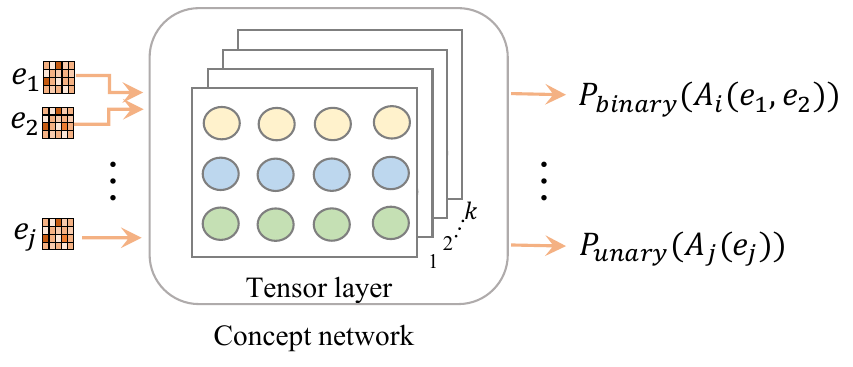}
   \caption{Concept network. The inputs are feature vectors of object pairs (e.g., $e_{1}$ and $e_2$) or objects (e.g., $e_j$), and outputs are probabilities of affiliation relationship labels (e.g., $P_{binary}(A_{i}(e_{1},e_{2}))$) or object labels (e.g., $P_{unary}(A_{j}(e_{j}))$). $k$ represents tensor layer and each layer is a predicate.}
   \label{fig:infer_net}
\end{figure}
\begin{figure*}[htpb]
\centering
\includegraphics[width=1\textwidth]{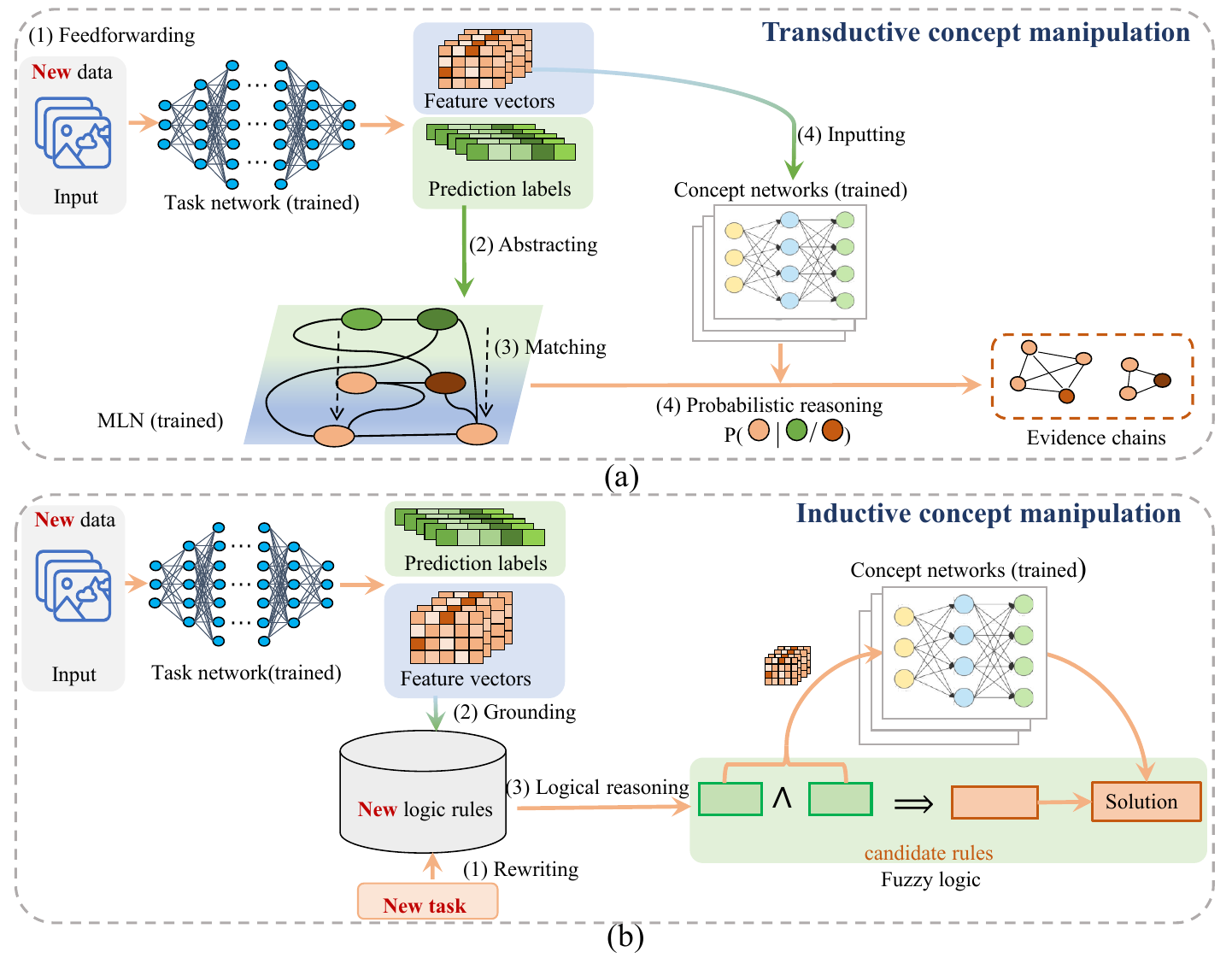}
\caption{Illustration of concept manipulation. (a) Transductive concept manipulation. The trained neural reasoning module predicts results, while the symbolic reasoning module provides interpretability.  (b) Inductive concept manipulation. The trained neural reasoning module generates feature vectors, which are used by the symbolic reasoning module for reasoning.}\label{concept_maniplation}
\end{figure*}

\subsubsection{Optimization}
The \texttt{NSF-SRL} model comprises two neural networks and a probabilistic graphical model, where the neural networks consist of a NRM and a concept network. The NRM is responsible for learning the features of concepts, while the concept network aims to infer the labels of query variables to approximate the posterior distribution. The symbolic reasoning module is responsible for learning a joint probability distribution to facillitate outcome inference.

The objective function 
$\log P_{\theta_1}$ of the neural reasoning module is typically differentiable and can be optimized using gradient descent. In this paper, the discrete logical knowledge within the symbolic reasoning module, represented as $\log P_{\theta_{2},w}$, is transformed into a probabilistic graphical form, making symbolic reasoning also differentiable through the introduction of a concept network for posterior inference. The model aims to minimize the objective function to facilitate end-to-end joint training of both modules. 
Specifically, during the E-step, the posterior distribution of the query variables is inferred, while in the M-step, the weights of the rules are learned. The training phase continues until the model reaches convergence. The parameters of the neural reasoning module, the concept network, and the symbolic reasoning module are denoted as $\theta_1$, $\theta_2$, and $w$, respectively.

To train the symbolic reasoning module, we need to maximize $O_{logic}$. However, the computation of the partition function  $Z(w)$ in $P_{\theta_2, w}(\hat{y},R)$ makes it intractable to optimize this objective function directly. Consequently, we introduce the variational EM algorithm and optimize the variational evidence lower bound (ELBO):

\begin{align}\label{eq:7}
\small
\begin{split}
     ELBO
    =E_{Q}[\log P_{\theta_2, w}(\hat{y},R)] -E_{Q}[\log Q(\hat{y}\mid R)],
\end{split}
\end{align}
where $Q(\hat{y}\vert R)$ is the variational posterior distribution. 

In general, we utilize the variational EM algorithm to optimize the ELBO. Specifically, we minimize the Kullback-Leibler (KL) divergence between the variational posterior distribution $Q(\hat{y}\vert R)$ and the true posterior distribution $P_w(\hat{y}\vert R)$ during the E-step. Due to the complex graphical structure among variables, the exact inference becomes computationally intractable. Therefore, we adopt a mean-field distribution to approximate the true posterior, inferring the variables independently as follows:
\begin{equation}\label{eq:8}
    Q(\hat{y}\vert R)=\prod_{ A_i\in A}Q(A_i).
\end{equation}

For computational convenience, traditional variational methods typically require a predefined distribution, such as the Dirichlet distribution, and then utilize traditional search algorithms to solve the problem. In contrast, we employ neural networks (concept networks in this paper) to parameterize the variational calculation in Eq. (\ref{eq:8}). Consequently, the variational process transforms into a parameter learning process for the neural networks. As illustrated in Fig. \ref{fig:infer_net}, the neural network is called the concept network and is used to compute the posterior $Q(A_i)$. Thus, $Q(A_i)$ is rewritten as $Q_{\theta_2}(A_i)$.

Based on the above analysis, combined with Eq. (\ref{eq:4}) and Eq. (\ref{eq:8}), Eq. (\ref{eq:7}) is rewritten as:
\begin{align}\label{eq:7_1}
  \begin{split}
    &ELBO= \sum_{r\in R}w_r\sum_{ a_r\in A_{r}}E_{Q_{\theta_2}}[\phi(a_{r})]-\log Z(w)\\
    &-E_{Q_{\theta_2}}[\sum_{A_i\in A}Q_{\theta_2}(A_i)]+\mathbb{C}.
    \end{split}
\end{align}

In Fig. \ref{fig:infer_net}, to attain predicate labels of the hidden variables, we first feed feature vectors into concept network, such as feature vector of an object pair ($e_1,e_2$) or the feature vector of a single object $e_j$. Then, the concept network outputs a binary predicate label if provided with feature vectors of an object pair; otherwise, it outputs a unary predicate label. For example, when we input the feature vector of an image of a zebra into the concept network, it can output the predicate ``zebra". 
Furthermore, to enhance the performance of the concept network through supervised information, we introduce a cross-entropy loss for optimization, which serves as a log-likelihood, 

\begin{equation}\label{eq:9}
\small
   O_{cro}=-\sum_{A_i\in A}Q_{\theta_2}(A_i)\log \hat{y_{i}}= -\log\Pi_{A_i\in A}\hat{y_{i}}^{Q_{\theta_2}(A_i)}.
\end{equation}

Thus, the overall E-step objective function becomes: 
\begin{equation}\label{eq:10}
\small
    O=\alpha O_{task} +\beta O_{logic}-\gamma  O_{cro},
\end{equation}
where $\alpha$, $\beta$ and $\gamma$ are the hyperparameter to control the weight. We maximize Eq. (\ref{eq:10}) to learn model parameters, the details are as follows: 

\begin{equation}\label{eq:parameter}
\small
    \{\theta_1^{\ast},\theta_2^{\ast}\}=\mathop{\arg\max}\limits_{\theta_1,\theta_2}O.
\end{equation}


In the M-step, the model learns the weights of the first-order logic rules. As we optimize these weights, the partition function $Z(w)$ in Eq. (\ref{eq:4}) is no longer constant, while $Q_{\theta2}$ remains fixed. The partition function $Z(w)$ consists of an exponential number of terms, rendering direct optimization of the ELBO intractable. To solve this issue, we employ pseudo-log-likelihood \cite{richardson2006markov} to approximate the ELBO, which is defined as follows:
\begin{equation}\label{eq:11}
\small
    P_{w}(\hat y,R) \simeq E_{Q_{\theta_2}}[\sum_{A_i\in A}\log P_{w}(A_i\vert {MB}_{A_i})],
\end{equation}
where ${MB}_{A_i}$ represents Markov blanket of the ground atom $A_i$. For each rule $r$ that connects $A_i$ to its Markov blanket, we optimize weights $w_r$ using gradient descent, and derivative is given by the following:
\begin{equation}\label{eq:12}
 \small
    \bigtriangledown_{w_r}E_{Q_{\theta_2}}[\log P_{w}(A_i\vert {MB}_{A_i})] \simeq \hat{y_{i}}- P_{w}(A_i\vert {MB}_{A_i}),
\end{equation}
where $\hat{y_{i}}$ = 0 or 1 if $A_i$ is an observed variable, and $\hat{y_{i}} = Q_{\theta_2}(A_i)$ otherwise.

\subsection{Concept Manipulation}
\label{sec4:3}
As mentioned in the overview of \texttt{NSF-SRL}, concept manipulation includes transductive and inductive concept manipulation methods. Consequently, we designed two corresponding approaches, as illustrated in Fig. \ref{concept_maniplation}. When the test data intersects with the training data, transductive concept manipulation employs the trained task network to predict results and utilizes probability graphical model to derive the FOLs corresponding to these predictions, providing explanations, as shown in Fig. \ref{concept_maniplation} (a). In contrast, when the test data is disjoint from the training data, inductive concept manipulation uses the trained task network to extract data features. By introducing new FOLs to generalize the model for addressing new tasks, fuzzy logic reasoning is then applied to deduce the prediction results, as depicted in Fig. \ref{concept_maniplation} (b).  

In the scenario depicted in Fig. \ref{concept_maniplation} (a), the categories of the training set and the test set overlap. As illustrated in Fig. \ref{overview}, the training data includes ``zebra", which is also present in the test data. The steps of transductive manipulation are as follows: (1) Feedforwarding: input new data and obtain prediction labels and features through the trained task network; (2) Abstracting: derive partial nodes as observed variables in the probabilistic graphical model from the predicted labels; (3) Matching: match these partial nodes with first-order logic rules in the Markov logic network to identify candidate rules; (4) Inputting: feed feature vectors into the concept network, retrieve the scores of the concepts, and apply probabilistic reasoning (Eq. (\ref{poster})) and fuzzy logic reasoning to obtain the probability score of each rule being true. Rules with high scores are selected as the evidence chain, interpreting the prediction labels. In this paper, we match the prediction results with ground atoms of the logic rules to achieve interpretability. A successful match indicates that the logic rules containing those ground atoms are triggered, and the corresponding clique composed of those nodes is selected. To quantify the likelihood that a candidate rule is true, we calculate the probability using t-norm fuzzy logic \cite{novak2012mathematical}. This process allows us to obtain evidence in the form of logic rules supporting the reasoning outcomes. To enhance interpretability, we select the most prominent piece of evidence in terms of a specific rule based on the posterior probability $P(r\mid \hat{y})$ as follows: 
\begin{equation}\label{poster}
P(r\mid \hat{y})=\prod_{A_i \in T_r}p({A_i}\mid \hat{y}),
\end{equation}
where $T_r$ is the candidate rule here. Here, $A_i$ is the ground atom sets in $T_r$. 

In the scenario depicted in Fig. \ref{concept_maniplation} (b), the categories of the training data and the test data do not overlap. As shown in Fig. \ref{overview}, the training data does not include ``tiger," whereas the test data does. Specifically, there are three steps for inductive manipulation: (1) Rewriting: rewriting logic rules based on the new task to accommodate specific requirements; (2) Grounding: grounding the logic rules using feature vectors from the task network; (3) Logic reasoning: inputting feature vectors of the concepts mentioned in the rule body of the candidate rules into the concept network to obtain the labels of the concepts. Subsequently, we reason the solution for the new task based on both the rule head and the rule body. This process can be seen as reprogramming for a new problem, utilizing the learned concepts from the previous step to tackle more complex problem scenarios. For instance, the model is trained on single-digit image addition and tested on multi-digit image addition tasks. By adopting this approach, the model can adapt its knowledge and reasoning capabilities to address new problems, thereby demonstrating the generalization capabilities of our method.

\section{Experiments}
In this section, we conduct experiments on various tasks, including supervised task (transductive concept manipulation), weakly supervised task (transductive concept manipulation and inductive concept manipulation), and zero-shot learning task (inductive concept manipulation), using classic datasets for validation. We first describe the datasets and evaluation metrics. Then, we report the empirical
results, including the performance, generalization, and interpretability across different tasks. Finally, we present ablation studies and hyperparameter analysis. The code is available at https://github.com/Dongranyu/NSF-SRL.

\subsection{Experimental Setup}
\textit{Tasks and datasets}: For the supervised task, we validate our approach on visual relationship detection task. The corresponding datasets are Visual Relationship Detection(VRD) \cite{lu2016visual} and VG200 \cite{xu2017scene}. For the weakly supervised task, we conduct experiments on a digit image addition task, utilizing the handwritten digit dataset MNIST. For the zero-shot learning task, we employ image classification for validation, using the AwA2 \cite{8413121} and CUB \cite{welinder2010caltech} datasets.

The \textbf{VRD} contains 5,000 images, with 4,000 images as training data and 1,000 images as testing data. There are 100 object classes and 70 predicates (relations). The VRD includes 37,993 relation annotations with 6,672 unique relations and 24.25 relationships per object category. This dataset contains 1,877 relationships in test set never occur in training set, thus allowing us to evaluate the generalization of our model in zero-shot prediction.

The \textbf{VG200} contains 150 object categories and 50 predicates. Each image has a scene graph of around 11.5 objects and 6.2 relationships. 70\% of the images is used for training and the remaining  30\% is used for testing.

The \textbf{MNIST} is a handwritten digit dataset and includes 0-9 digit images. In this paper, the task is to learn the “single-digit addition” formula given two MNIST images and a ``addition”
label. To implement the experiment on single-digit image addition, we randomly choose the initial feature of two digits to concat a tuple and take their addition as their labels. MNIST has 60,000 train sets and 10,000 test sets.

The \textbf{AwA2} consists of 50 animal classes with 37,322 images. Training data contains 40 classes with 30,337 images, and test data has 10 classes with 6,985 images. Additionally, AwA2 provides 85 numeric attribute values for each class.

The \textbf{CUB} comprises 11,788 images spanning 200 bird classes, each associated with 312 attributes. Among these classes, 150 classes are designated as seen during training, while the remaining 50 are unseen and used for evaluation.

The \textbf{logic rules}.
In this paper, logic rules encode relationships between a subject and multiple objects for visual relationship detection. Here, we build logic rules in an artificial way for VRD and VG200 datsets. That is, we take relationship annotations together with their subjects and objects to construct a logic rule according to the annotation file in the dataset. For example, we can obtain a logic rule as $\textsf{laptop}(x) \wedge \textsf{next to}(x,y) \Rightarrow \textsf{keyboard}(y) \vee \textsf{mouse}(y)$ by the above method. As a result, the numbers of logic rules are 1,642. Unlike VRD datasets, MNIST has no relationship annotation. To adapt to our weakly supervised task, we define corresponding logic rules, e.g., combining two single-digit labels and their addition label as logic rule. For example, $\textsf{digit} (x, d_{1}) \wedge \textsf{digit} (y, d_{2}) \Rightarrow \textsf{addition} (d_{1}+d_{2}, z)$, where the rule head is the addition label, and the rule body is two single-digit labels. In zero-shot image classification, we design logic rules for the AwA2 and CUB datasets, where the rule head is animal categories and the rule body consists of their attributes. For instance, $\textsf{catlike} (x) \wedge \textsf{tawny} (x) \wedge \textsf{spot}(x) \Rightarrow \textsf{leopard} (x)$.

\textit{Metrics}:
For VRD, we adopt evaluation metrics same as \cite{zhang2019large}, which runs \textbf{Relationship detection (ReD)} and \textbf{Phrase detection (PhD)} and shows recall rates (Recall$@$) for the top 50 /100 results, with $k=1,70$ candidate relations per relationship proposal (or $k$ relationship predictions for per object box pair) before taking the top 50/100 results. \textbf{ReD} is inputting an image and outputting labels of triples and boxes of the objects. \textbf{PhD} is inputting an image and output labels and boxes of triples.

For VG200, we use the same evaluation metrics used in \cite{zhang2019large}, including 1) \textbf{Scene Graph Classification (SGCLS)}, which is to predict labels of the subject, object, and predicate given ground truth subject and object boxes; 2) \textbf{Predicate Classification(PCLS)}, where predict predicate labels are given ground truth subject and object boxes and labels. Recall$@$ under the top 20/50/100 predictions are reported.

For MNIST, AwA2 and CUB, we adopt accuracy(\textbf{Acc}) to evaluate the performance of the model. They are defined as Eq. (\ref{eq:ACC}).
\begin{align}\label{eq:ACC}
\small
\begin{split}
   Acc= \frac{TP + TN}{TP+TN+FP+FN},
\end{split}
\end{align}
where $TP$ denotes true positive, $TN$ denotes true negative, $FP$ indicates false positive, and $FN$ is false negative.

For the logic rule, we compute the probability of a logic rule that is true as an evaluation of logic rules. Here, we adopt Łukaseiwicz of t-norm fuzzy logic \cite{novak2012mathematical}.

\subsection{Digit Image Addition Task}
In the context of neural-symbolic studies, digit image addition serves as a benchmark task, and MNIST dataset is recognized as a benchmark dataset. We evaluate the performance of our \texttt{NSF-SRL} model by comparing it against several neural-symbolic approaches and convolutional neural networks (CNNs). The neural-symbolic approaches considered include DeepPSL \cite{dasaratha2023deeppsl}, DeepProbLog \cite{manhaeve2021neural}, and NeurASP \cite{yang2021neurasp}. This paper assesses the model's performance specifically on the single-digit image addition task, where two single-digit images are input into \texttt{NSF-SRL}, and the output is the predicted addition result. Furthermore, to verify the model's generalization capability, we also perform the multi-digit image addition task in Section 5.5.

In the digit image addition task, the neural reasoning module first extracts image features using a CNN. These features are then processed through two fully connected layers to produce a 10-dimensional output vector. The activation functions employed in this neural network structure are ReLU and Softmax. We set the learning rate to 1e-4 and train the model for 15,000 epochs. Additionally, we utilize a batch size of 64 during training and a batch size of 1,000 during testing.

\begin{figure*}[htpb]
\centering
\includegraphics[width=1\linewidth]{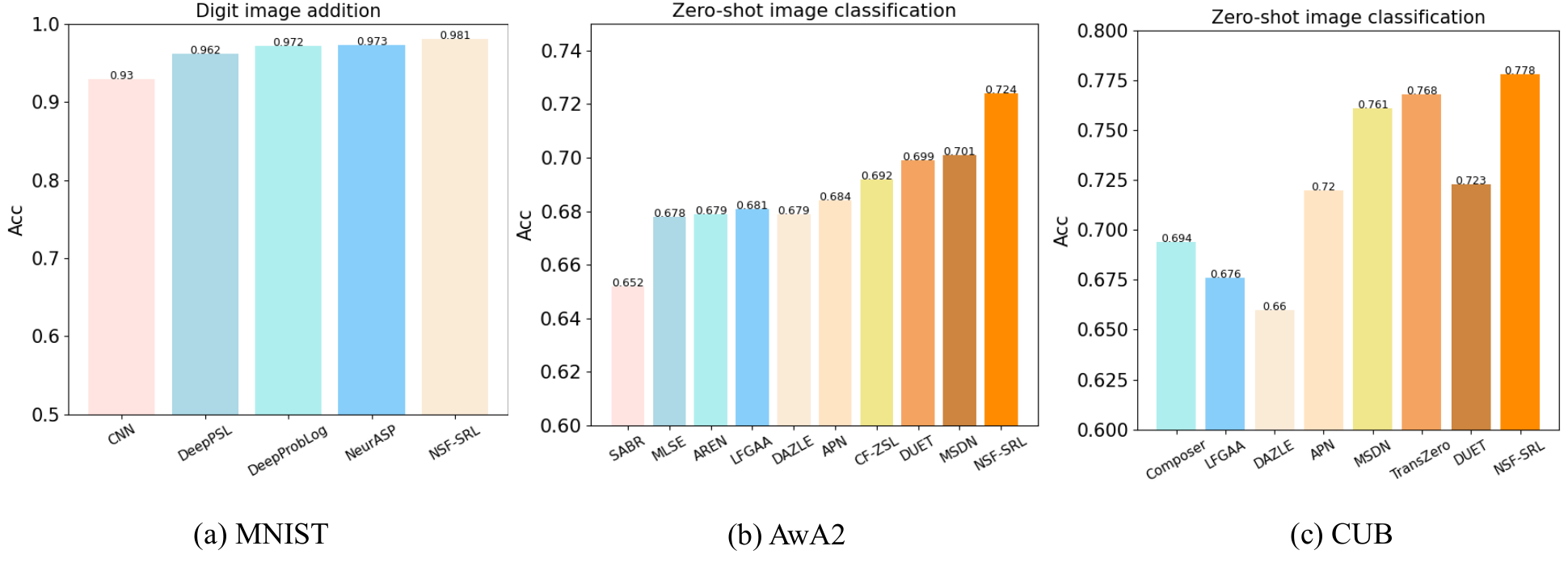}
\caption{Performance of \texttt{NSF-SRL} and comparison methods on digit image addition and zero-shot image classification tasks: (a) MNIST ; (b) AwA2 ; (c) CUB.} \label{performance}
\end{figure*}

Fig. \ref{performance} (a)  presents the results of \texttt{NSF-SRL} alongside comparison methods for the digit image addition task. By comparing the performance of \texttt{NSF-SRL} with that of the other methods, we observe that \texttt{NSF-SRL} achieves decent performance. This finding underscores the feasibility of \texttt{NSF-SRL} in circumventing the reliance on strong supervised information typically required in conventional deep learning approaches. By integrating symbolic knowledge, \texttt{NSF-SRL} effectively leverages additional supervisory signals, such as data labels and relationships between data, resulting in improved model performance.

\subsection{Zero-shot Image Classification Task}
In contrast to the digit image addition task, the zero-shot image classification task is inherently more complex. This task involves training a model on images of seen classes, enabling it to recognize images of unseen classes. The objective of the neural reasoning module in this context is to learn a mapping function from the visual space to the semantic space, thereby extracting image features of the objects. The symbolic reasoning module first receives these image features from the neural reasoning module, then models the logic rules using a MLN to learn the joint probability distribution. Finally, it employs a concept network to calculate the posterior probability of the joint distribution, predicting attribute labels and combining these labels according to the established logic rules. 

\begin{figure}[htpb]
\centering
\includegraphics[width=1\linewidth]{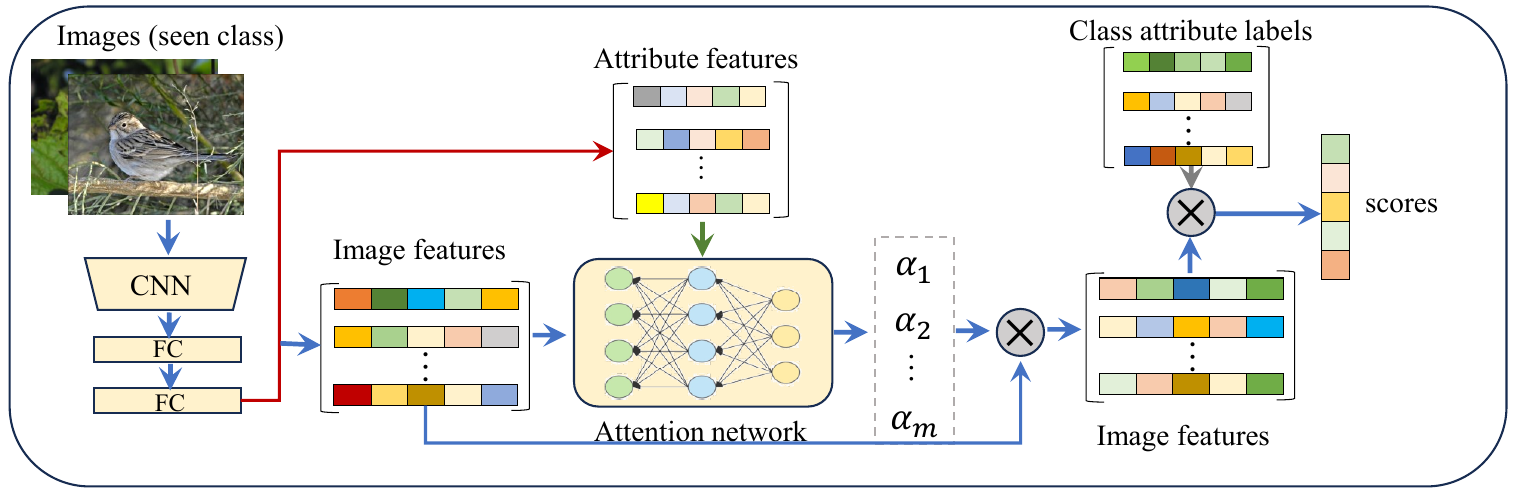}
\caption{The neural reasoning module on zero-shot image classification.} \label{image_classification}
\end{figure}

In the zero-shot image classification task, the neural reasoning module is a CNN initialized with a pre-trained GoogleNet. Given an input image, we first use the CNN to extract initial visual features. These features are then fed into an attention network to attain discriminative
image features. To enhance data augmentation, images undergo random cropping before being input into the model. For optimization, we employ the Adam optimizer with the following configurations: 15 epochs, a batch size of 64, and a learning rate of 1e-4. The specific
neural architecture is illustrate in the Fig.\ref{image_classification}.

The symbolic reasoning module is implemented as a MLN, which integrates the neural reasoning module with FOL to extract discriminative image features. Additionally, this module enables the trained model to adapt from recognizing seen classes to unseen classes. Specifically, the symbolic reasoning module employs the MLN to learn the joint probability distribution between symbolic discriminative features and classes, predicting the labels of these features by calculating the posterior probability. Consequently, the symbolic reasoning module effectively combines the image features extracted by the neural reasoning module with FOL to perform fuzzy logic reasoning and derive class labels. The introduction of the MLN provides an efficient method for integrating visual features and symbolic discriminative features, thereby enhancing the model's generalization capability to unseen classes. This joint modeling approach captures the associations between image features and attributes, ultimately improving the model's performance in zero-shot image classification.

Zero-shot image classification is a complex reasoning task that current neural-symbolic methods struggle to address effectively. Consequently, we primarily adopted deep learning-based contrastive approaches. Fig. \ref{performance} (b) presents the results for the AwA2 dataset, comparing our method against baseline methods such as SABR \cite{paul2019semantically}, MLSE \cite{ding2019marginalized}, AREN \cite{xie2019attentive}, LFGAA \cite{liu2019attribute}, DAZLE \cite{huynh2020fine}, APN \cite{xu2020attribute}, CF-ZSL \cite{Yang2021collaborative}, DUET \cite{chen2023duet}, and MSDN \cite{chen2022msdn}. Fig. \ref{performance} (c) presents the comparative results on the CUB dataset. The methods included in this comparison are Composer \cite{huynh2020compositional}, LFGAA \cite{liu2019attribute}, DAZLE \cite{huynh2020fine}, APN \cite{xu2020attribute}, MSDN \cite{chen2022msdn}, TransZero \cite{chen2022transzero}, and DUET \cite{chen2023duet}. 

From Fig. \ref{performance}, it is evident that our \texttt{NSF-SRL} achieves optimal performance across different datasets, validating the effectiveness of the model. This success can be attributed to the logical rules that model the relationships between attribute features, seen categories, and unseen categories, including co-occurrence relationships. Such rules facilitate the model in capturing these relationships, thereby enhancing classification performance. Additionally, this experiment highlights that incorporating symbolic reasoning with FOL enhances the robustness of the model.

\subsection{Visual Relationship Detection Task}
Visual relationship detection, similar to zero-shot image classification, is a complex task that aims to identify objects within an image and the relationships between them. These relationships can be represented as triplets (subject, predicate, object). In this context, the neural reasoning module serves as a deep learning-based specifically designed for visual relationship detection, extracting label concepts of both objects and their relationships from the input image. Conversely, the symbolic reasoning module functions as a two-layer probabilistic graphical model, intended to integrate the learned object and relationship labels while guiding the learning process of the visual reasoning module.

\begin{figure}[htpb]
\centering
\includegraphics[width=1\linewidth]{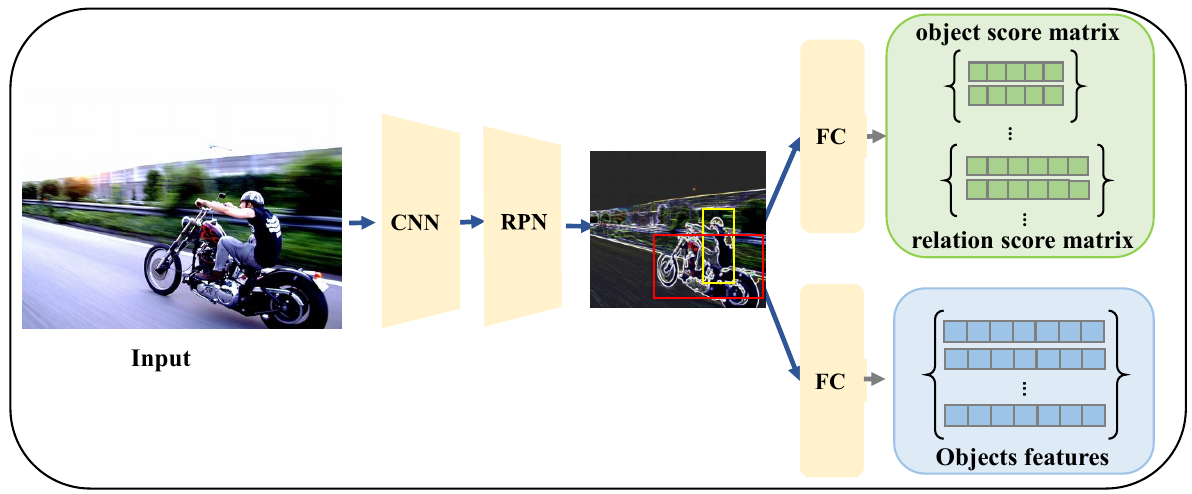}
\caption{The neural reasoning module on visual relationship detection task.} \label{VRD}
\end{figure}

For the visual relationship detection task, our neural reasoning module is based on the architecture described in \cite{zhang2019large}. It consists of two components: a visual module and a semantic module. The visual module primarily extracts visual features using a CNN, specifically employing layers conv1\_1 to conv5\_3 of VGG16 to generate a global feature map of the image. Subsequently, the subject, relation, and object features are region-of-interest (ROI) pooled and processed through two fully connected layers to produce three intermediate hidden features. The semantic module, on the other hand, processes word vectors corresponding to the subject, relation, and object labels via a multilayer perceptron (MLP) to generate embeddings. Before training, we initialize each branch using pre-trained weights from the COCO dataset \cite{lin2014microsoft} and adopt word2vec \cite{mikolov2013distributed} for the word vectors in our experiments. Specifically, we train our model for 7 epochs with a learning rate set to 1e-4, and the dimension of the object feature is established at 512. The specific neural architecture is illustrated in Fig. \ref{VRD}.

\begin{figure}[htpb]
\centering
\includegraphics[width=1\linewidth]{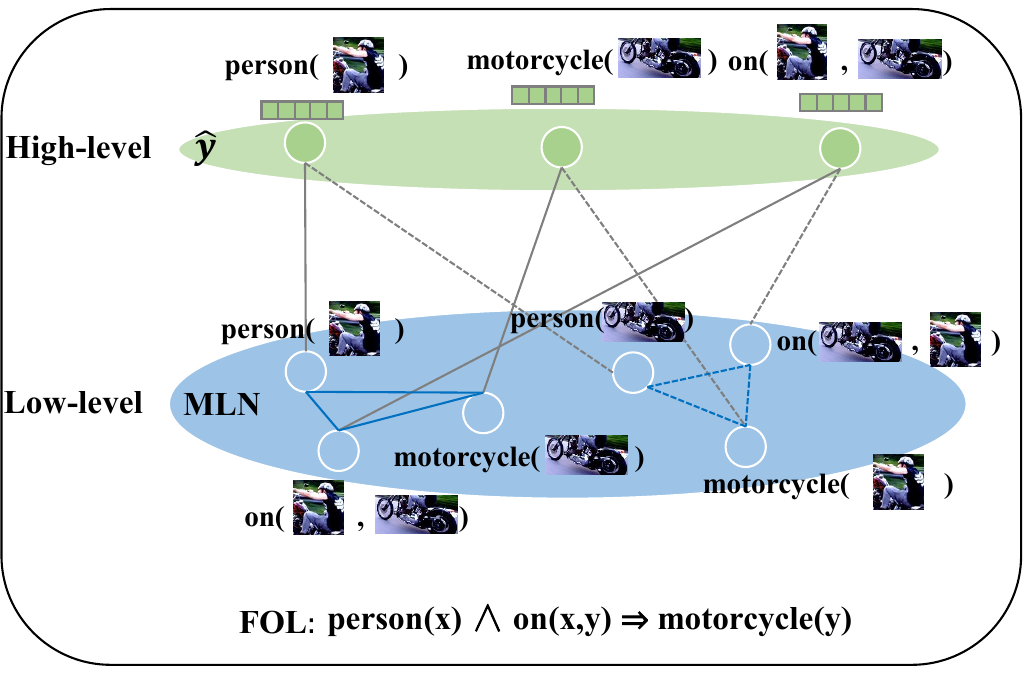}
\caption{The symbolic reasoning module on visual relationship detection task.} \label{vrd_symbolic}
\end{figure}

As illustrated in the Fig. \ref{vrd_symbolic}, the symbolic reasoning module is structured as a bi-level probabilistic graphical model, where the high-level layer represents the prediction results (pseudo-labels) generated by the neural reasoning module. In contrast, the low-level layer consists of the ground atoms of MLN. This module consists of two types of nodes (random variables) and cliques (potential functions): the prediction labels from the neural reasoning module in the high-level layer nodes and the ground atoms of the MLN in the low-level layer nodes. Let $\hat y = \{ \hat{y_{1}}, \hat{y_{2}},...\}$ denote the set of high-level nodes (pseudo-labels), and let $A = \{A_{1}, A_{2},...\}$ represent the set of low-level nodes, comprising the ground atoms in the FOLs. A clique $\{ \hat{y_i}, A_j\}$ signifies the correlation between these levels, while another clique $A_{r}$ represents the ground atoms of a FOL. Consequently, the custom term $\mathbb{C}$ can be defined as $\sum_{\hat{y_i} \in \hat{y},A_j \in A}  \phi_1(\hat{y_i},A_j)$ in Eq. (\ref{eq:4}).

Existing neural-symbolic methods, such as DeepProbLog, have not been validated on complex tasks like visual relationship detection. Therefore, our comparative methods are restricted to those based solely on deep learning. The experimental results of \texttt{NSF-SRL} and several comparative methods are presented in Table \ref{tab:1} for VRD dataset. As not all comparative methods specified $k$ in their experiment, we report results as ``$free\ k$" when treating $k$ as a hyperparameter. The results indicate that our \texttt{NSF-SRL} outperforms the comparative methods in most cases. The enhancements offered by the symbolic reasoning module can be attributed to two key factors.  First, the symbolic reasoning module is structured as a probabilistic graphical model that effectively captures dependencies between variables, facilitating a more accurate modeling of complex relationships. Second, our logic rules are constructed based on the co-occurrence relationships among predicates, suggesting that when one object is present, another is likely to appear as well. By maximizing the joint probability of the probabilistic graphical model, we effectively enhance the co-occurrence probability during the training phase.

Table \ref{tab:2} presents the results on the VG200 dataset. Notably, the state-of-the-art methods do not specify a clear value for $k$ in this context.  Therefore, we report the performance of our \texttt{NSF-SRL} model with $k$ = 1. Our results demonstrate that \texttt{NSF-SRL} outperforms existing methods across two metrics in Recall$@$20/50/100, highlighting the advantages of leveraging symbolic knowledge through logic rules. Furthermore, while PCLS emphasizes relationship recognition, \texttt{NSF-SRL} achieves a superior score on the PCLS evaluation metric, indicating that the incorporation of logic rules enhances relationship recognition capabilities within the model.

\begin{table*}[htbp]
\scriptsize
\renewcommand{\arraystretch}{1.3}
\centering
\caption{Test performance of visual relationship detection. The recall results for the top 50/100 in ``ReD” and ``PhD” are reported, respectively. 
The best result is highlighted in bold. ``$-$'' denotes the corresponding result is not provided.}
\label{tab:1}
\resizebox{\linewidth}{!}{
\begin{tabular}{c|c|c|c|c|c|c|c|c|c|c|c|c}
\hline
\rowcolor{gray!10}\textbf{Methods}& \multicolumn{2}{|c|}{\textbf{ReD}}& \multicolumn{2}{|c|}{\textbf{PhD}}& \multicolumn{4}{|c|}{\textbf{ReD}} & \multicolumn{4}{|c}{\textbf{PhD}}\\
\hline
\rowcolor{gray!10} & \multicolumn{4}{|c|}{$free\ k$}& \multicolumn{2}{|c|}{$k=1$}& \multicolumn{2}{|c|}{$k=70$}& \multicolumn{2}{|c|}{$k=1$}& \multicolumn{2}{|c}{$k=70$}
\\
\hline
\rowcolor{gray!10} Recall$@$ &50& 100& 50& 100& 50& 100& 50& 100& 50& 100& 50& 100\\
\hline
Lk distilation\cite{yu2017visual}& 22.7& 31.9& 26.5& 29.8& 19.2& 21.3& 22.7& 31.9& 23.1& 24.0& 26.3&29.4 \\

Zoom-Net\cite{yin2018zoom}& 21.4& 27.3& 29.1& 37.3& 18.9& 21.4&21.4& 27.3& 28.8& 28.1&29.1&37.3 \\

CAI+SCA-M\cite{yin2018zoom}& 22.3& 28.5& 29.6& 38.4& 19.5& 22.4&22.3& 28.5& 25.2& 28.9&29.6& 38.4\\

MF-URLN\cite{zhan2019exploring}& 23.9& 26.8& 31.5& 36.1& 23.9& 26.8&$-$&$-$ & 23.9 &26.8 &$-$ &$-$ \\
LS-VRU\cite{zhang2019large}& 27.0& 32.6& 32.9& 39.6& 23.7& 26.7& 27.0& 32.6& 28.9& 32.9&32.9& 39.6\\

GPS-Net\cite{lin2020gps}& 27.8& 31.7& 33.8& 39.2&$-$ & $-$& 27.8& 31.7& $-$& $-$& 33.8& 39.2\\

UVTransE\cite{hung2020contextual}& 27.4& 34.6& 31.8& 40.4& 25.7& \textbf{29.7}&  27.3& 34.1& 30.0& 36.2& 31.5& 39.8\\
NMP\cite{hu2022neural}& 21.5& 27.5& $-$& $-$&20.2 & 24.0& 21.5& 27.5& $-$& $-$& $-$& $-$\\
\hline
\texttt{NSF-SRL} & \textbf{29.4}&\textbf{35.3} &\textbf{36.2} &\textbf{43.0} & \textbf{26.2}& 29.4& \textbf{29.4}& \textbf{35.3}& \textbf{32.3}& \textbf{36.4}&\textbf{36.2}& \textbf{43.0}\\
\hline
\end{tabular}}
\end{table*}

\subsection{Generalization}
Evaluating a model's generalization ability is essential, as it reflects its adaptability and robustness across diverse scenarios. In this study, generalization refers to the model's predictive performance on unseen samples. For example, the model is initially trained on a
single-digit image addition task and subsequently tested on a multi-digit
image addition task. Zero-shot image classification serves as an experiment that validates the model's generalization capabilities. Consequently, we only focus our  experimental validation on visual relationship detection and digit image addition tasks.
\subsubsection{Visual Relationship Detection}
We evaluated the performance of our \texttt{NSF-SRL} model against the baseline LS-VRU in a zero-shot learning scenario. In this context, the training and testing data comprise disjoint sets of relationships from the VRD dataset, as illustrated in Fig. \ref{generalization} (a). The results demonstrate that \texttt{NSF-SRL} outperforms LS-VRU across various recall metrics, highlighting LS-VRU's limitations in handling sparse relationships. In contrast, \texttt{NSF-SRL} effectively incorporates symbolic knowledge and language priors, making it less susceptible to the challenges posed by sparse relationships.

\subsubsection{Digit image Addition}
\begin{figure}[htpb]
\centering
\includegraphics[width=1\linewidth]{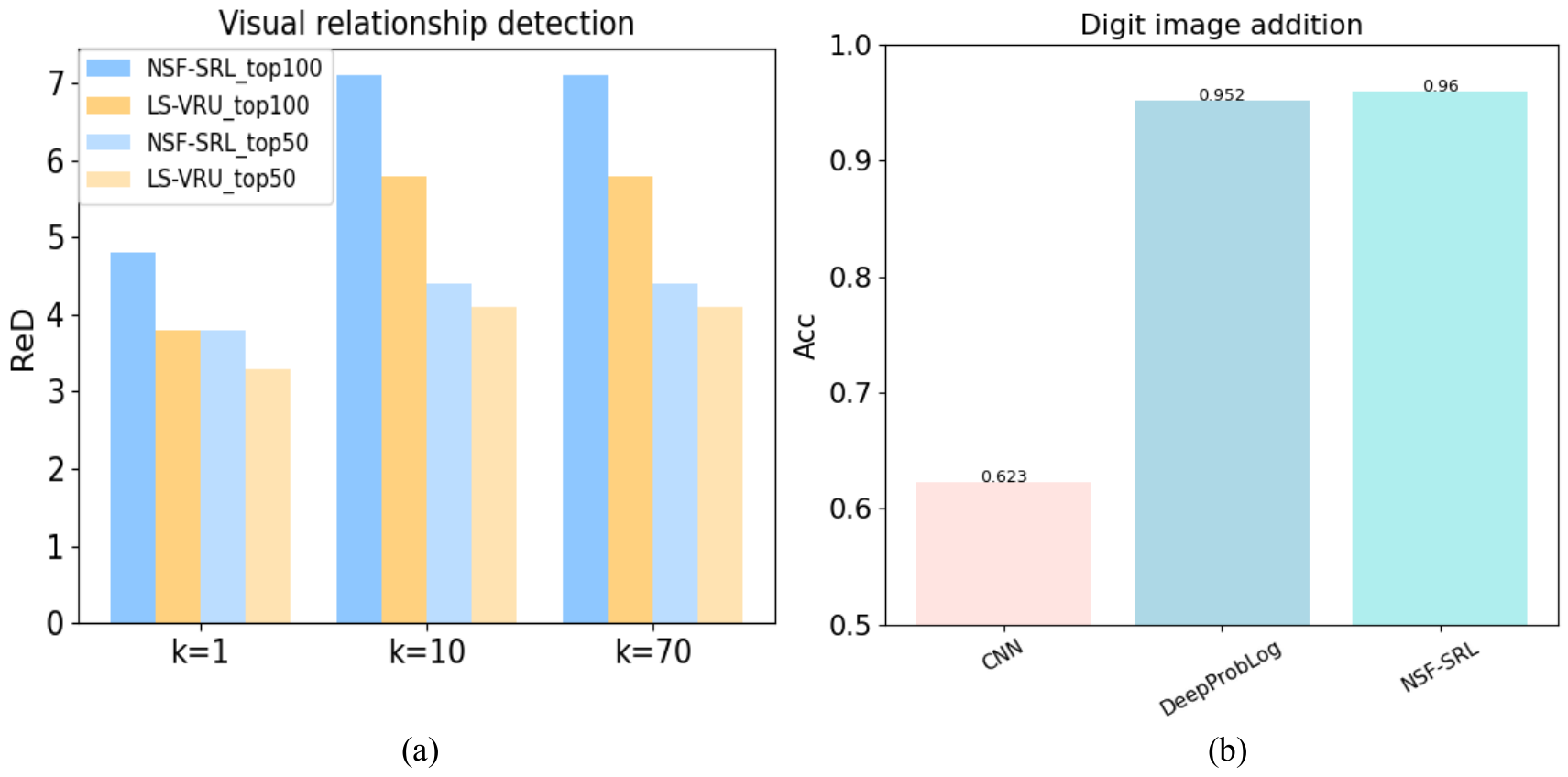}
\caption{Generalization of \texttt{NSF-SRL} and comparison methods on visual relationship detection and digit image addition tasks. (a) Visual relationship detection.  Larger ReD indicates better results. (b) Multi-digit image addition.}\label{generalization}
\end{figure}

We validate the generalization capability of \texttt{NSF-SRL} in multi-digit task by comparing it to the baseline. In multi-digit image addition, the input consists of two lists of images, each representing a digit, with each list corresponding to a multi-digit number. The label reflects the sum of these two numbers. In our experiment, a CNN is trained on the multi-digit image addition dataset to test the multi-digit image addition task, while we apply the learned model from the single-digit image addition task to this scenario. As shown in Fig. \ref{generalization} (b),  the results illustrate the enhanced prediction accuracy in the multi-digit image addition task by leveraging concepts acquired during the single-digit task.Our findings indicate a significant improvement compared to other methods, underscoring the flexibility of our model, which can generalize from simpler tasks to more complex ones by adapting its logic rules. Notably, this generalization is facilitated by the shared learnable concepts between the two tasks.

\begin{table}[htpb]
\renewcommand{\arraystretch}{1.3}
\scriptsize
\centering
\caption{Comparative results for top 50/100 in ``SGCLS” and ``PCLS” respectively on the VG200 dataset. The best result is highlighted in bold.} 
\label{tab:2}
\begin{tabular}{c|c|c|c|c|c|c}
  \hline
\diagbox{Recall$@$}{{Metrics}}& \multicolumn{3}{|c|}{\textbf{SGCLS}}& \multicolumn{3}{|c}{\textbf{PCLS}}\\
\hline
Methods&  20& 50& 100 &  20& 50& 100
\\
\hline
VRD\cite{lu2016visual}& $-$&11.8 &14.1 &$-$ &27.9&35.0\\

Ass-Embedding\cite{newell2017pixels}&18.2 &21.8 &22.6 &47.9 &54.1&55.4\\

Mess-Passing\cite{xu2017scene}&31.7 &34.6 &35.4 &52.7 &59.3&61.3\\

Graph-RCNN\cite{yang2018graph}&$-$ &29.6 &31.6 &$-$ &54.2&59.1\\

Per-Invariant\cite{herzig2018mapping}&$-$&36.5 &38.8 &$-$ &65.1& 66.9\\

Motifnet\cite{zellers2018neural}&32.9 &35.8&36.5 &58.5 &65.2&67.1 \\

LS-VRU\cite{zhang2019large}&36.0&36.7&36.7&66.8&68.4&68.4\\

GPS-Net\cite{lin2020gps}& 36.1&39.2&\textbf{40.1}&60.7&66.9 &68.8 \\
\hline
\texttt{NSF-SRL}($k=1$)&\textbf{37.0} &\textbf{39.3}&39.3&\textbf{67.8} &\textbf{69.1}& \textbf{70.0}\\
\hline
\end{tabular}
\end{table}

\begin{figure*}[htpb]
\centering
\includegraphics[width=\textwidth]{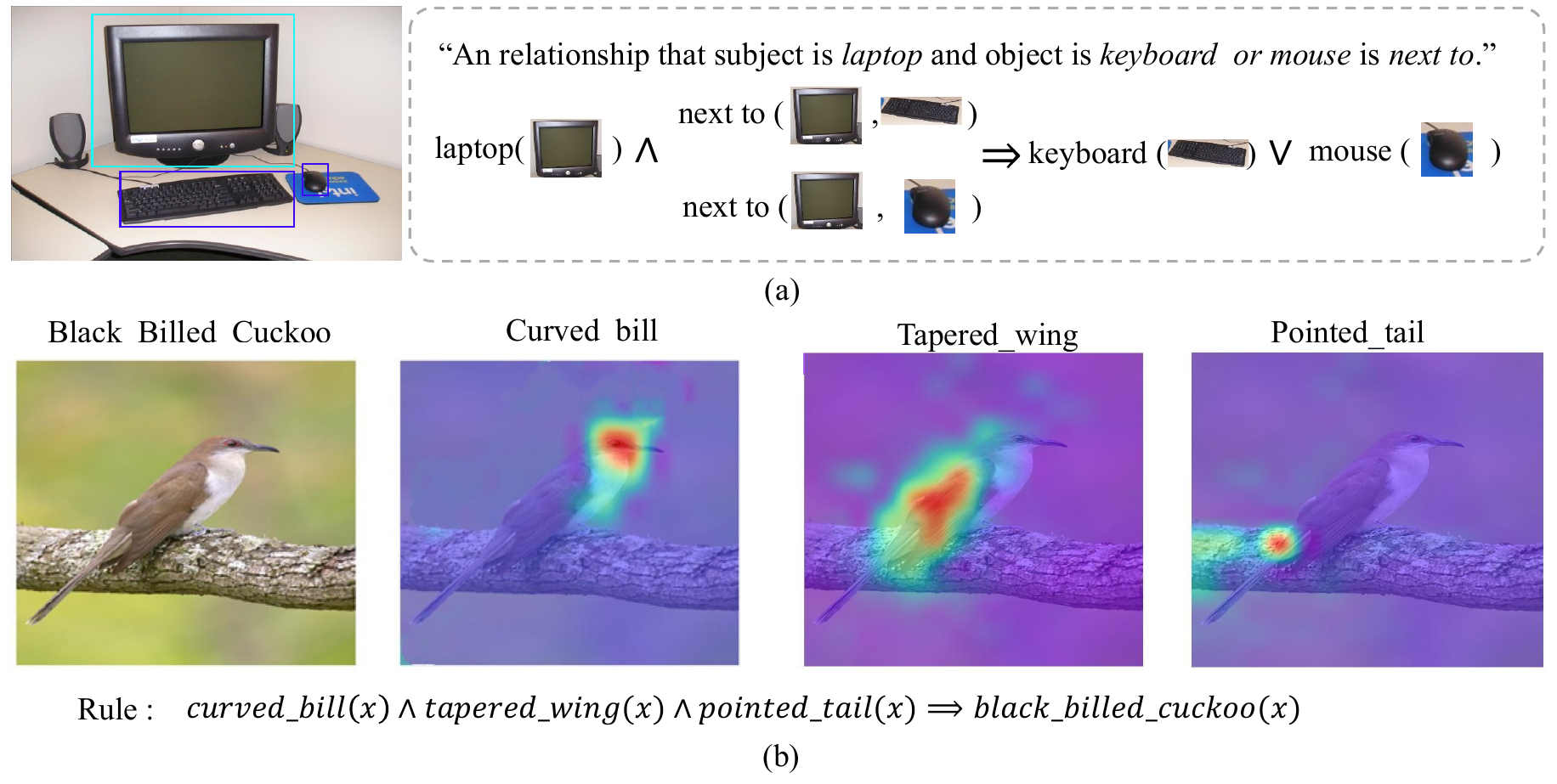}
\caption{Interpretability analysis. (a) An example illustrating the interpretability of \texttt{NSF-SRL}.  
For example, why is the relationship ``next to” detected between a ``laptop” and a ``keyboard” or ``mouse” in an image? According to Eq. (\ref{poster}), the model identifies the most confident logic rule: $\textsf{laptop}(x) \wedge \textsf{next to}(x,y) \Rightarrow \textsf{keyboard}(y) \vee \textsf{mouse}(y)$. This demonstrates that the reasoning results of \texttt{NSF-SRL} align with common sense. (b) Visualization of the learned discriminative image features by our model. Key features, such as the shape of the bill, wing, and tail, are highlighted, providing a visual explanation of the model's reasoning.}\label{interpretability}
\end{figure*}

\subsection{Interpretibility}
We employ visual relationship detection and zero-shot image classification tasks to demonstrate the interpretability of our results.
In the context of visual relationship detection, Fig. \ref{interpretability} (a) illustrates the reasoning behind the identified relationship ``next to" between
``laptop" and either the ``keyboard" or ``mouse". According to Eq. (\ref{poster}), when the subject is a ``laptop" and the object is either ``keyboard" or ``mouse", the relationship ``next to" is assigned the highest confidence by the logic rule $\textsf{laptop}(x) \wedge \textsf{next to}(x,y) \Rightarrow \textsf{keyboard}(y) \vee \textsf{mouse}(y)$. 

In zero-shot image classification, we 
used heatmaps to visualize the discriminative image features. As shown in Fig. \ref{interpretability} (b), the highlighted regions represent the discriminative features captured by our model. By combining the predicted
discriminative feature labels with the logic rules, the model can infer class labels. This transparent reasoning process facilitates easy understanding of the model’s decision-making when presented with an image. For instance, when the model identifies an image as $black\_billed\_vuckoo$, it justifies its prediction by highlighting features such as a $curved\_bill$, $tapered\_wing$ and $pointed\_tail$ in the image, and logically deduces that the object possessing these features belongs to the $black\_billed\_vuckoo$ class, based on the applied rule.

\begin{figure}[htpb]
\centering
\includegraphics[width=1\linewidth]{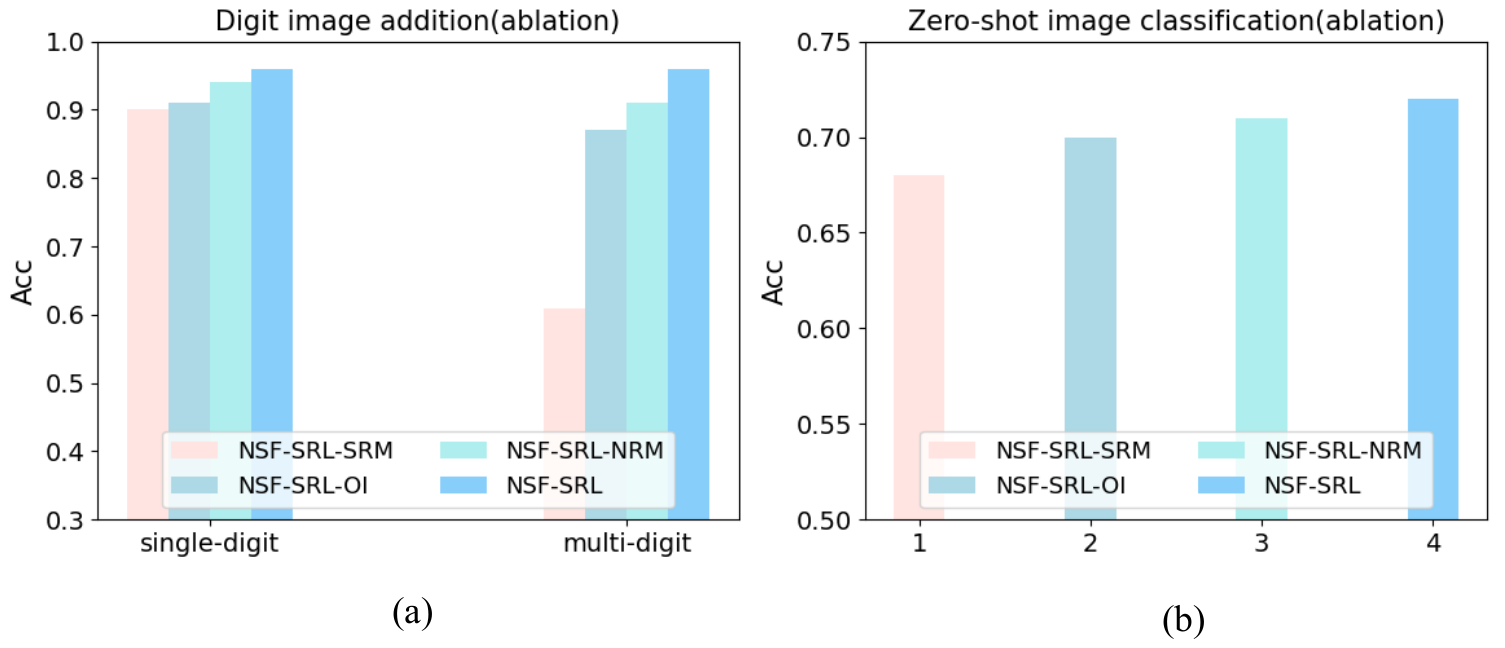}
\caption{Ablation results on digit image addition and zero-shot image classification tasks. }\label{ablation}
\end{figure}

\subsection{Ablation Studies}
During the training phase, we conduct an extensive analysis of various factors that may affect downstream task performance. These factors include the hyperparameters $\alpha$, $\beta$, $\gamma$. This comprehensive evaluation framework provides deeper insights into the influence of these factors on model performance.

To investigate the impact of model trade-offs on reasoning, we designed three variants to assess the effect of individual components on \texttt{NSF-SRL}. Specifically, we derived these variants from the optimized objective in Eq. (\ref{eq:10}) by adjusting the values of the trade-off factors. The three variants are as follows: (1) \texttt{NSF-SRL-SRM} ($\alpha = 1, \beta = 0, \gamma = 0$): excluding the symbolic reasoning module, (2) \texttt{NSF-SRL-NRM} ($\alpha = 1/2, \beta = 1, \gamma = 1$): reducing the visual reasoning module by half, and (3) \texttt{NSF-SRL-OI} ($\alpha = 1, \beta = 1, \gamma = 0$): omitting the cross-entropy of observed variables. We conducted experiments on digit image addition and zero-shot image classification tasks to evaluate performance of \texttt{NSF-SRL} and its variants. The results are presented in Fig. \ref{ablation} (a) and Fig. \ref{ablation} (b), respectively. 

In Fig. \ref{ablation} (a), we observe that the performance of the \texttt{NSF-SRL-NRM} variant is higher compared to its \texttt{NSF-SRL-SRM} counterparts. This indicates that the symbolic reasoning module is crucial in weakly supervised tasks. This is likely due to the limited availability of supervised information in such tasks. Specifically, in weakly supervised tasks, the input images are not individually labeled but only labeled by the addition task. As a result, the NRM module may have a more restricted role in these tasks. Moreover, this finding highlights the importance of incorporating symbolic knowledge.

In Fig. \ref{ablation} (b), we observe that the correlations among the components of SRM, VRM, and OI have a significantly positive impact on zero-shot image classification. Furthermore, the performance of our model is notably enhanced when SRM is applied, confirming the effectiveness of the symbolic knowledge integrated into the model. We conclude that symbolic knowledge helps the model adapt to new environments, specifically in recognizing unseen classes.

\subsection{Hyperparameter Analysis}
To analyze the robustness of our \texttt{NSF-SRL} framework and determine optimal hyperparameters, we conducted extensive experiments to evaluate the effects of epoch settings and loss weights (in Eq. (\ref{eq:10})).

\begin{figure}[htpb]
\centering
\includegraphics[width=1\linewidth]{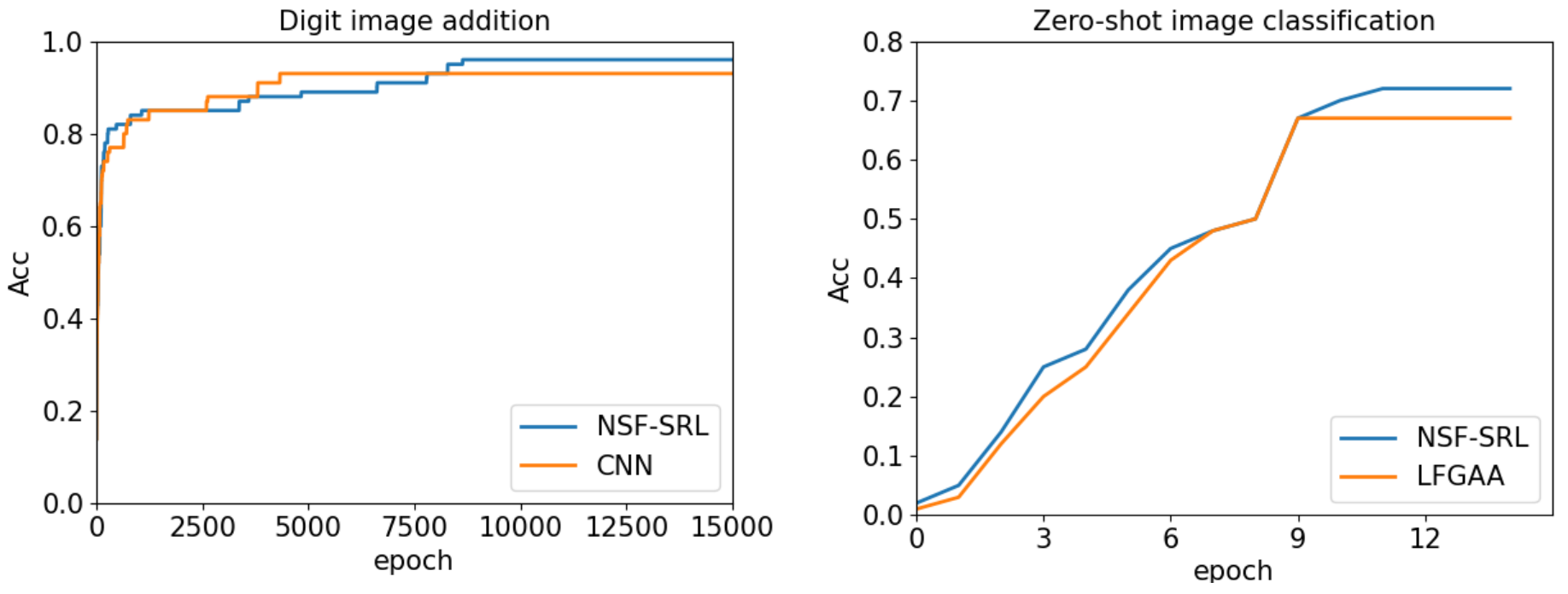}
\caption{Effects of different epochs for the \texttt{NSF-SRL} on digit image addition and zero-shot image classification.}\label{epoch}
\end{figure}

1) Effects of Epoch: In Fig. \ref{epoch}, we present the fine-tuning results for models trained with varying numbers epochs, evaluated based on
accuracy (Acc) for both digit image addition and zero-shot image classification tasks. The figures clearly show that both \texttt{NSF-SRL} and the baseline models exhibit an upward trend as the number of iterations increases. This trend suggests that the models continue to benefit from longer training, indicating that extended training can further improve performance until convergence. Additionally, the baseline models converge faster than \texttt{NSF-SRL}, which may be due to differences in model architecture, such as CNN or LFGAA having fewer parameters to learn.

\begin{figure}[htpb]
\centering
\includegraphics[width=1\linewidth]{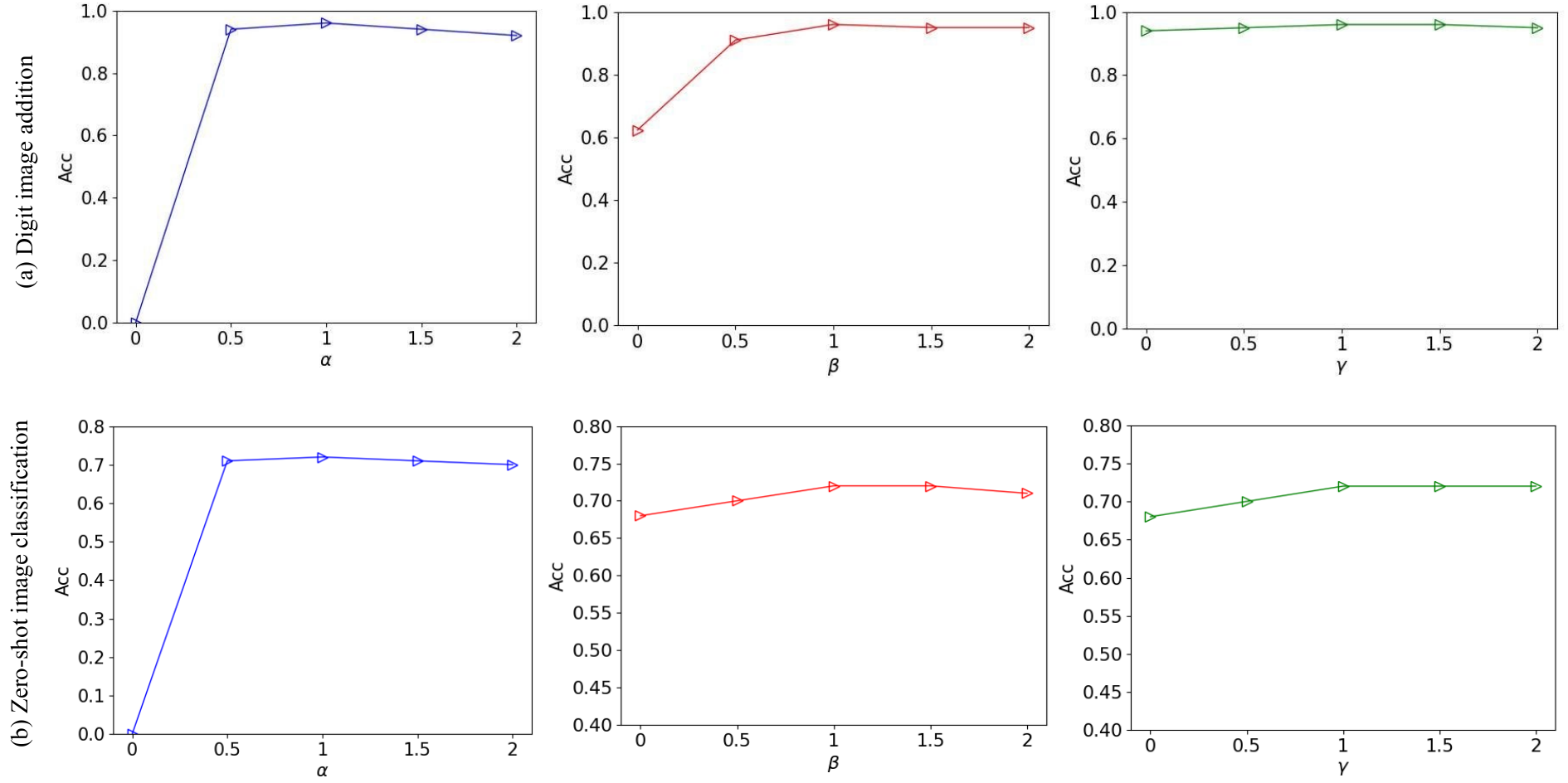}
\caption{Effects of loss weights that control their corresponding loss terms on digit image addition and zero-shot image classification tasks, i.e., $\alpha$, $\beta$ and $\gamma$.} \label{digit_hyperparameter}
\end{figure}

2) Effects of Loss Weights: 
In this section, we analyze the impact of the loss weights $\alpha$, $\beta$ and $\gamma$ on their respective loss terms. We experimented with a range of values \{0, 0.5, 1, 1.5, 2\} for these weights across digit image addition and zero-shot image classification tasks. The results are illustrated in Fig. \ref{digit_hyperparameter}. When $0<\alpha < 0.5$, all evaluation metrics exhibit an upward trend, while for $\alpha > 0.5$, the performance across all evaluation strategies remains consistent. Additionally, \texttt{NSF-SRL} demonstrates relative insensitivity to $\beta$ and $\gamma$ when set to larger values (e.g., greater than 0.5). Based on these observations, we set $\alpha$, $\beta$, and $\gamma$ to 1, 1, and 1, respectively, in our experiments.

\section{Conclusion}\label{conclusion}
In this study, we introduce \texttt{NSF-SRL}, a general model in neural-symbolic systems. Our goal is to improve the model's performance and generalization, while also providing interpretability of the results.   Additionally, we propose a novel evaluation metric to quantify the interpretability of the deep model. Our experimental results demonstrate that \texttt{NSF-SRL} outperforms state-of-the-art methods across various reasoning tasks, including supervised, weakly supervised, and zero-shot image classification scenarios, in terms of both performance and generalization. Furthermore, we highlight the interpretability of \texttt{NSF-SRL} by providing visualizations that clarify the model’s reasoning process. 

In practice, the \texttt{NSF-SRL} can find applications in diverse scenarios beyond the experimental tasks discussed in this paper. For instance, in healthcare, the model can be leveraged for medical image analysis and patient diagnosis. By amalgamating symbolic reasoning with deep learning capabilities, it can assist physicians in disease diagnosis and treatment planning while enhancing diagnostic reliability through interpretability. In the financial sector, the \texttt{NSF-SRL} can be instrumental in fraud detection and risk assessment by effectively managing complex data patterns with its hybrid approach.

In our \texttt{NSF-SRL} framework, the manual definition of logic rules may restrict the breadth of acquired rule knowledge and involves labor costs. On top of this foundational work, a potential enhancement would involve enabling the model to autonomously learn rules from data, leading to a more efficient and adaptive system.


%



\ifCLASSOPTIONcompsoc
  \section*{Acknowledgments}
\else
  \section*{Acknowledgment}
\fi

This work was supported by the National Key R\&D Program of China under Grant Nos. 2021ZD0112500;  the National Natural Science Foundation of China under Grant Nos. U22A2098, 62172185, 62202200, and 62206105.

\ifCLASSOPTIONcaptionsoff
  \newpage
\fi

\bibliographystyle{IEEEtran}
\bibliography{reference}

\vfill


\end{document}